\def\year{2022}\relax
\documentclass[letterpaper]{article} 
\usepackage{aaai22}  
\usepackage{times}  
\usepackage{helvet}  
\usepackage{courier}  
\usepackage[hyphens]{url}  
\usepackage{graphicx} 
\usepackage{natbib}  
\usepackage{caption} 
\DeclareCaptionStyle{ruled}{labelfont=normalfont,labelsep=colon,strut=off} 
\usepackage{algorithm}
\usepackage{algorithmic}
\usepackage{multirow}

\usepackage{newfloat}
\usepackage{listings}
\usepackage{booktabs}
\floatstyle{ruled}
\newfloat{listing}{tb}{lst}{}
\floatname{listing}{Listing}
\newcommand{\argmin}{\mathop{\mathrm{argmin}}\limits}

\title{Rethinking Neural Networks With Benford's Law}
\author {
    Surya Kant Sahu,\textsuperscript{\rm 1}
    Abhinav Java, \textsuperscript{\rm 2}\textsuperscript{\rm 1}
    Arshad Shaikh \textsuperscript{\rm 1} and 
    Yannic Kilcher \textsuperscript{\rm 3}
}
\affiliations {
    \textsuperscript{\rm 1} The Learning Machines\\
    \textsuperscript{\rm 2} Delhi Technological University\\
    \textsuperscript{\rm 3} ETH Zürich \\
    surya.oju@pm.me, java.abhinav99@gmail.com
}

\usepackage{bibentry}

\begin{document}

\maketitle

\begin{abstract}
Benford's Law (BL) or the Significant Digit Law defines the probability distribution of the first digit of numerical values in a data sample. This Law is observed in many naturally occurring datasets. It can be seen as a measure of naturalness of a given distribution and finds its application in areas like anomaly and fraud detection. In this work, we address the following question: Is the distribution of the Neural Network parameters related to the network's generalization capability? To that end, we first define a metric, MLH (Model Enthalpy), that measures the closeness of a set of numbers to Benford's Law and we show empirically that it is a strong predictor of Validation Accuracy. Second, we use MLH as an alternative to Validation Accuracy for Early Stopping, removing the need for a Validation set. We provide experimental evidence that even if the optimal size of the validation set is known beforehand, the peak test accuracy attained is lower than not using a validation set at all. 
Finally, we investigate the connection of BL to Free Energy Principle and First Law of Thermodynamics, showing that MLH is a component of the internal energy of the learning system and optimization as an analogy to minimizing the total energy to attain equilibrium.
\end{abstract}

\section{Introduction}
\label{introduction}

Benford's Law (BL) has been observed in many naturally occurring populations, including the physical constants, populations of countries, areas of lakes, stock market indices, tax accounts, etc. \cite{physics}. Researchers have also discovered the presence of this law in natural sciences \cite{sambridge2010benford}, image gradient magnitude \cite{jolion2001images}, synthetic and natural images \cite{acebo2005benford}, etc. Attempts have been made to explain the underlying reason for BL’s emergence for specific domains. However, a universally accepted explanation does not yet exist.

The fact that BL occurs in many naturally occurring datasets, and the samples which don't not obey BL are probable anomalies, is one of the reasons why BL is also known as \textit{"The Law of Anomalous Numbers"}. Due to this, BL has been used to ascertain fraud in taxing and accounting and in machine learning literature for Anomaly Detection, such as detecting GAN-generated images \cite{ganbenford}.

Through this work, we hope to bring attention to the Machine Learning community that BL can have potential applications in training and evaluation of Neural Networks. We summarize our contributions as follows:
\begin{itemize}
    \item We show with strong empirical evidence that a metric that we propose based on BL contains non-trivial information about an NN's generalization to unseen data.
    \item We present a direct application of this result, by replacing Validation metrics for early stopping, and hence removing the need of a validation set.
    \item We connect our results to the Free Energy Principle \cite{pmlr-v119-gao20a}, \cite{alemi2018therml} and \cite{friston2010free} and attempt to explain why BL contains information about an NN's generalization. 
    \item And hence define a cheap-to-compute Information Criterion for a given network.
\end{itemize}

\section{Preliminaries}
\subsection{Thermodynamics of Machine Learning}
Previous work has established the formal connection of Thermodynamics and machine learning. \cite{alemi2018therml} define four information-theoretic functionals, out of which, we focus on Relative Entropy $S$. It measures the entropy between the distribution $p(\theta|X, Y)$ that is assigned after training on data $(X, Y)$ prior $q(\theta)$ for model parameters $\theta$. $S$ can measure the risk of overfitting the parameters.
\begin{equation}
    S \equiv \log\frac{p(\theta | X, Y)}{q(\theta)}
    \label{eq:rel-entropy}
\end{equation}

As the authors claim, this measure is intractable.

\subsection{Free-Energy Principle and Information Criteria}
\paragraph{Free-Energy Principle} \cite{friston2010free} is a well-known principle that tries to explain the mechanism of learning and behaviour in living beings (referred to as "agents"). This principle states that agents take actions to sensory input, and its own internal state through an internal model of the world. This model is updated based on the outcome of the action. The learning objective, according to the Free-Energy Principle, is to minimize "surprise" in addition to minimizing the complexity of the learned model.

\begin{figure*}[htb]
\centering
{\includegraphics[width=0.90\textwidth]{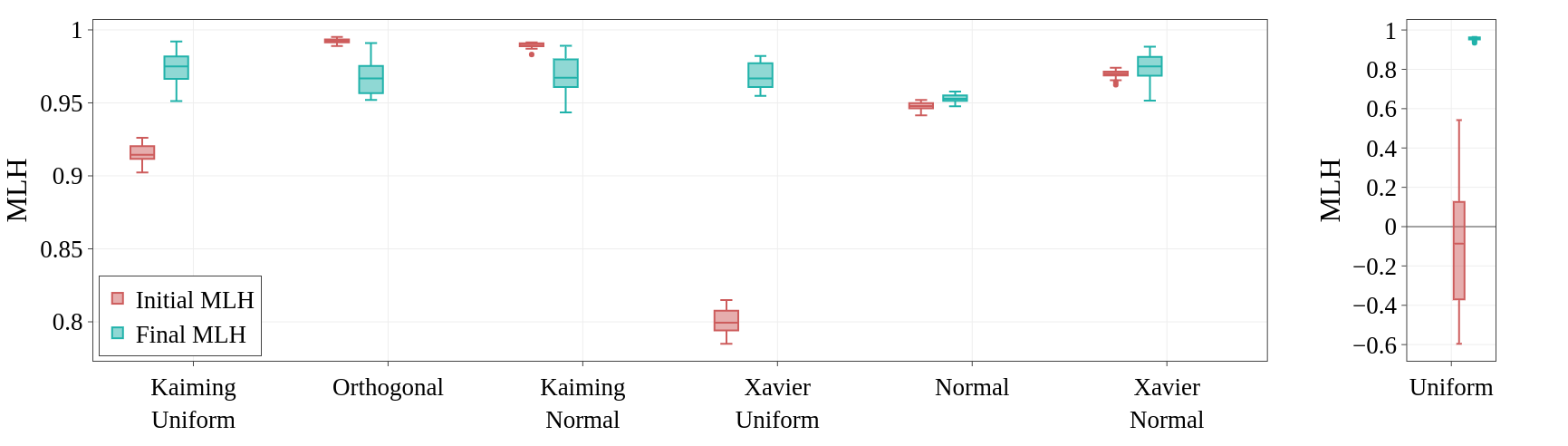}}
\caption{MLH for Trained and Randomly-Initialized Models. Regardless of initialization, the MLH of trained models remains high.}
\label{fig:neural_nets_follow_bl}
\end{figure*}

This statement can easily be applied to Machine Learning and Bayesian Inference; the minimization objective is then given as:
\begin{equation}
    \argmin_{\theta} J(\theta) = E_{x\in X}[L(x; \theta)] + D(\theta) 
    \label{eq:free-energy}
\end{equation}

where $J(\theta)$ is the energy that is to be minimized, the first term measures the error for a dataset $X$, and the second term measures the complexity of the model with parameters $\theta$.

Simply put, the learned model must have low prediction error while also not being complex. Note that this statement also relates to Occam's Razor and Information-based Criteria in ML literature.

\paragraph{Information-based Criteria} are used frequently for model selection in the ML Community. Bayesian Information Criterion (BIC) and Akaike Information Criteria (AIC) \cite{Burnham2003ModelSA} are some of the widely-used criteria for model selection. 
\begin{equation}
    AIC(m) = -2 \log L(m) + 2 p(m) 
\end{equation}
\begin{equation}
    BIC(m) = -2 \log L(m) + 2 p(m) \log n 
\end{equation}
where $m$ is a model, $L(m)$ is the error of the model $m$, $p(m)$ is the number of parameters of $m$, and $n$ is the number of data-points used to learn $m$.

It can be clearly seen that AIC and BIC are special cases of \ref{eq:free-energy}, where number of parameters is used as the measure for model complexity $D$.

\section{Neural Nets and Benford's Law} \label{NNfollowsBL}

\begin{figure}[tb]
    \centering
    \includegraphics[width=0.95\columnwidth]{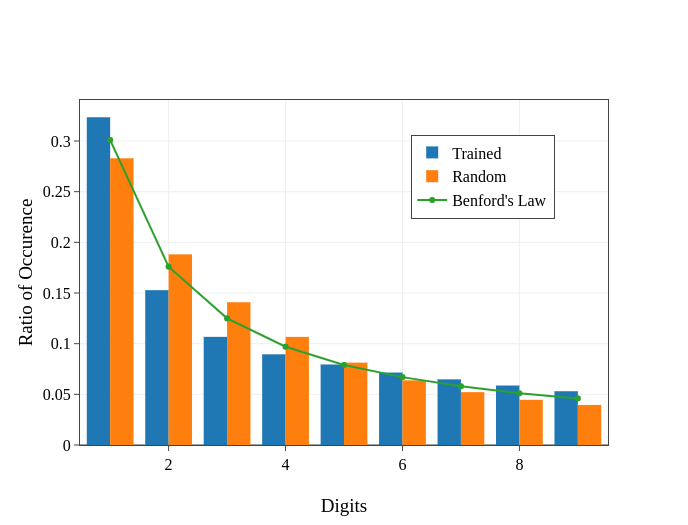}
    \caption{ResNet152; Trained and Random Weights' Significant Digit Distribution v/s. Benford's Law. Note that the trained model follows BL more closely. }\label{fig:bl-resnet}
\end{figure}

BL defines a probability distribution of a given sample's significant (leftmost) digit. The leftmost non-zero digit’s occurrence in the observations of a population is log-uniform for several datasets, with {1} occurring the maximum number of times, followed by {2, 3, till 9}. According to Benford's Law (BL) \cite{benfords}, the probability for a sample having a significant digit $d$ is given as follows:
\begin{equation}
    P_B = P(d) = log_{10}(d + \frac{1}{d}), d = 1, 2, 3, ..., 9 \label{eq:benford}
\end{equation}

As it is known that RGB images' pixel values follow BL, we hypothesize that Neural Network weights might also follow BL, for which, we devise a simple metric to measure the similarity of histograms of significant digits of model parameters. We define a simple metric MLH, that measures the correlation between Benford's Law and histogram of significant digits of a given set.

MLH is based on the Pearson's Correlation Coefficient \cite{pearson1895vii} is defined as follows:
    \begin{equation}
        MLH(\theta) = PearsonR(BinCount(\theta), P_B) \label{eq:MLH}
    \end{equation}
    \begin{equation}
        BinCount(\theta) = \frac{[f_0, f_1, ..., f_9]}{D_{\theta}} \label{eq:bincount}
    \end{equation}
    Here, $BinCount(\theta)$ is the distribution of Significant Digits of network parameter set $\theta$. $P_B$ is the distribution defined by BL, $f_k$ is is the frequency of significant digit $k$ occurring in $\theta$, $D_{\theta}$ is the dimensionality of $\theta$. 
    
    We did not include parameters that are initialized with a constant value, such as Bias and BatchNorm parameters. In our implementation, we multiply all elements in the set by a constant $10^{10}$ so that the resultant elements are greater than zero, and then take the first non-zero digit. This representation is required for a fast vectorized\footnote{Code is provided in the appendix} implementation of  $BinCount(.)$. Note that multiplying with a constant scalar doesn't change the distribution of significant digits due to BL's property of \textit{Scale Invariance} \cite{hill1995base}. 

This formulation of MLH allows simple implementation and interpretation of the values: The higher the MLH value, the higher the parameters' match with BL\footnote{Substituting with JS-Divergence yields mirrored curves.}.

\begin{figure*}[htb]
    \centering
    \includegraphics[width=\textwidth]{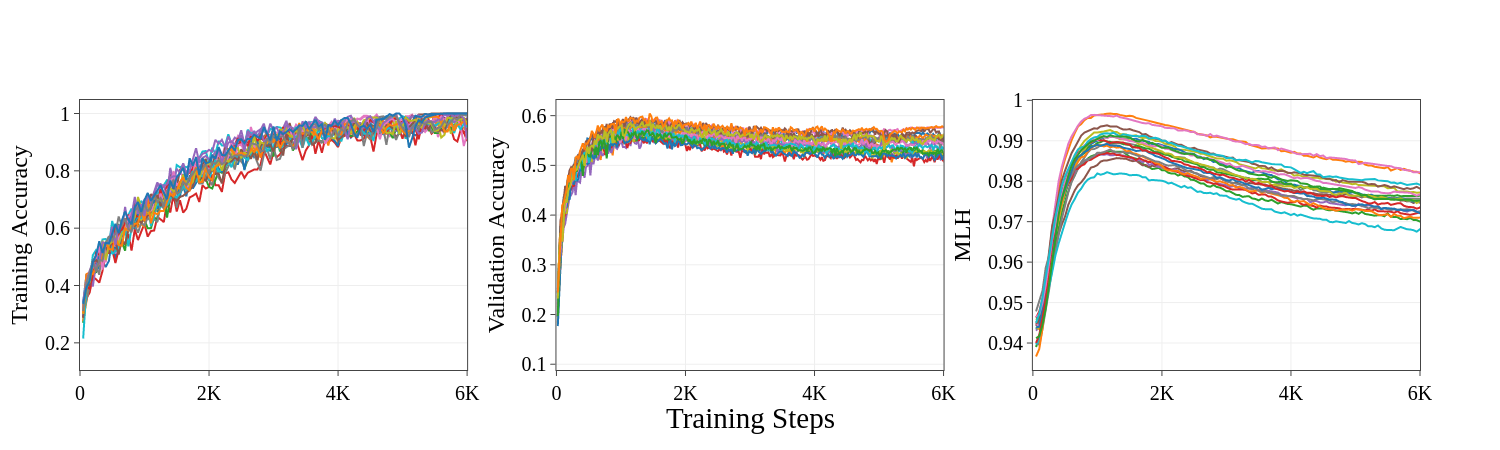}
    \caption{AlexNet-like models without dropout trained on CIFAR10. (Left to Right) Training accuracy, validation accuracy and MLH against training iterations. At around 1K iterations, the validation accuracy drops, while the training accuracy reaches 1. It can be clearly observed that the proposed metric MLH follows a similar trajectory to the validation accuracy. }
    \label{fig:runs}
\end{figure*}

In Fig.\ref{fig:bl-resnet},  we compare the Mantissa Distribution of a ResNet152 trained on ImageNet and a randomly initialized ResNet152 (Xavier Normal; Bias and BatchNorm are initialized with a constant). We see that both closely follow BL ($>$0.99 Pearson's R). 

\subsection{Effect of Initialization on MLH} \ref{sec:init}
This result points to the possibility that the way Neural Network weights (Xavier Normal) are initialized follow BL closely at the initialization, and after training, the closeness increases even further. 

In the this section, we present results on initializing networks with various methods, and show that they achieve high MLH at the end of training regardless of how the network was initialized. To show that a Neural Net's match with BL is not just due to the virtue of the initialization method. In Fig. \ref{fig:neural_nets_follow_bl}, we show results of an experiment where we repeatedly train $20$ Shallow Alexnet-like networks each on CIFAR10 for each initialization method, including initializing sub-optimally; we show that the MLH after training is always high regardless of the initial MLH\footnote{Default PyTorch values were used for Initialization.}

In the Appendix we explore how MLH varies throughout the depth of pretrained Transformers and ImageNet models, and a peculiar pattern is observed; attempt to find the effect of training data on MLH for various architectures (LSTMs, CNNs, MLPs etc.).

\section{Enthalpy Information Criterion (EIC)}

We can think of BL as a prior over significant digits of parameters $\theta$. The choice of BL as a prior is substantiated by the fact that $MLH$ is not an artifact of initialization \label{sec:init} and therefore possibly contains non-trivial information about a neural network's initialization. In Eq. \ref{eq:rel-entropy}, $S$ measures the entropy between the distribution over parameters and the prior, however MLH measures distribution of significant digits' closeness to BL. If we assume that the prior distribution $q(\theta)$ to be approximated by Benford's Law; $p(\theta)$ by $Bincount(\theta)$, we can see that $S$ is approximated by $MLH(\theta)$. Estimating prior distribution $q(\theta)$ over the parameter set would otherwise be a tedious exercise.

We now define a novel Information Criterion based on MLH:
\begin{equation}
    EIC(\theta) = - A_{\theta} - MLH_{\theta} 
    \label{eq:eic}
\end{equation}

It can be observed that in Fig. \ref{fig:bl-resnet}, even the randomly initialized network has MLH value over $0.99$. In experiments, we found that the mean MLH of all runs across all steps in RGB\footnote{MLH is lower for MNIST and FMNIST.} Datasets (CIFAR, Stanford Dogs, Oxford Flowers etc.) to be $0.974$, while the minimum and maximum were $0.9462$ and $0.9999$ respectively. We min-max scale MLH to bring it to the range $[0, 1]$.
\begin{equation}
    EIC_{scaled}(\theta) = - A_{\theta} - \frac{MLH_{\theta} - 0.9462}{0.0537} 
    \label{eq:eic-scaled}
\end{equation}

EIC is related to other information criteria such as Bayesian Information Criteria (BIC) \cite{10.1214/aos/1176344136} or Akaike Information Criteria (AIC) \cite{Akaike1998}. However, BIC and AIC both use the number of parameters of the model as a measure of model complexity; unlike EIC, which computes a statistic based on the values of the parameters, hence model complexity can differ for the same model with different parameter values. 

We show in the following sections that the property of $S$ to measure the risk of overfitting, is demonstrated by MLH and consequently by EIC.

\begin{figure*}[htb]
    \centering
    \includegraphics[width=0.65\textwidth]{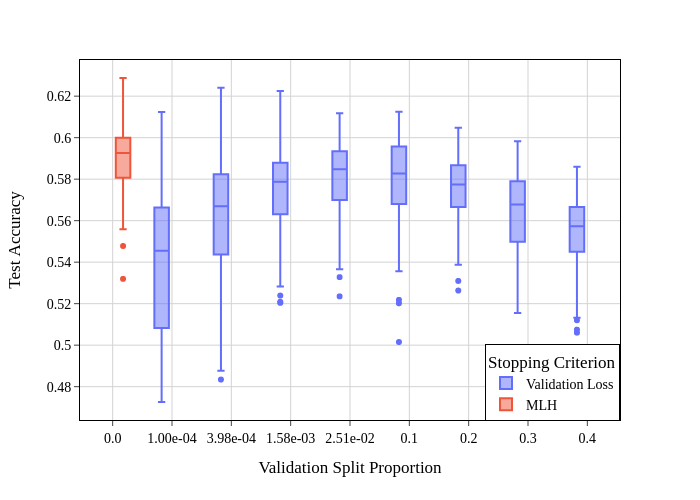}
    \caption{(Red) Test accuracy of models trained using MLH as early stopping. (Blue) Test accuracy of validation proportions used to dictate early stopping.  Complements Table \ref{tab:valprop}.}\label{fig:valprop}
\end{figure*}

\section{MLH and Validation Accuracy}\label{sec:valacc}

\begin{table}
\centering
\begin{tabular}{l|c} 
\toprule
{\textbf{Method / Metric}} & {\textbf{Spearman's R}}  \\ 
\hline
A                                             & 0.214                                      \\
MLH                                           & 0.583                                      \\
-1 * EIC w/o Scaling                          & 0.292                                      \\
\textbf{-1 * EIC w/ Scaling}                  & \textbf{0.654}                             \\
\textbf{-1 * EIC - SR}                        & \textbf{0.679}                             \\
GPR(A)                                        & 0.418                                      \\
GPR(MLH)                                      & 0.627                                      \\
\textbf{GPR(MLH, A)}                          & \textbf{0.774}                             \\
\bottomrule
\end{tabular}
\caption{Correlation between proposed metrics and Validation Accuracy. A is the Training accuracy. GPR($x_1, x_2, x_3, \dots$) refer to GaussianProcessRegressor fitted with $x_1, x_2, x_3, ...$ as input features. It can be seen that Min-max scaling drastically increases correlation for EIC, and improves beyond the correlation of MLH only.} \label{tab:corrval} 
\end{table}

For deep learning projects, researchers and practitioners split the available data into at least three sets: Training, Validation, and Test.

Practitioners use the validation set metrics for many purposes, one such use case is deciding when to stop the training i.e. detect overfitting and stopping the training before the model overfits to the train set. This is known as Early Stopping. Early Stopping requires a criterion that can determine the degree of overfitting, usually given by the validation set accuracy (as a proxy for generalization to unseen data).

In this section, we explore whether closeness to BL is related to validation accuracy. To this end, we train 100 shallow AlexNet-like models without dropout on the CIFAR10 dataset, manually split into train (90\%) and validation (10\%) sets; and collect MLH, validation and train accuracy over the course of training.

In Fig. \ref{fig:runs}, we randomly select a few training runs and plot their metrics. We observe that MLH and validation accuracy follow a similar trajectory. We make this observation concrete by computing Spearman's correlation coefficient \cite{spear} between various metrics and validation accuracy. Spearman's Correlation Coefficient measures the monotonic relationship between two random samples.

In Table \ref{tab:corrval}, we observe a strong correlation\footnote{\textit{p}-values are omitted because their values were extremely low (order of $10^{-12}$).} between MLH and validation accuracy. The correlation is even higher if we define a quantity that simply sums training accuracy and min-max scaled MLH. This result shows that MLH and training accuracy could be used to estimate validation accuracy. In section \ref{sec:thermodynamics}, we present the motivation behind using both MLH and training accuracy for estimating validation accuracy.
    
We also fit Gaussian Process Regression (GPR) models from scikit-learn \cite{scikit-learn} on the metrics collected so that it learns a function mapping from either training accuracy or MLH or both to validation accuracy. We observe that the GPR which uses both training accuracy and MLH has the highest correlation with validation accuracy, strongly indicating that MLH contains non-trivial information about the network's generalization performance.

Similarly, we also run Symbolic Regression using gp-learn to learn a mapping from Training Accuracy and MLH to Validation Accuracy. We obtain a simple program that doesn't require min-max scaling while achieving higher correlation than Eq. \ref{eq:eic-scaled}.

\begin{equation}
    EIC_{sr}(\theta) = -\frac{\log MLH(\theta)}{A(\theta)}
    \label{eq:eic-sr}
\end{equation}

\section{Early Stopping with MLH}\label{sec:mlh-stopping}

In this section, we present a direct application of MLH. In the previous section, we established that MLH is strongly correlated to validation accuracy. We use this result to replace validation set-based criteria for Early Stopping, while using the data saved as additional training data.

\begin{table}[htb]
\centering
\begin{tabular}{c|c|c|c} 
\toprule
\begin{tabular}[c]{@{}l@{}}\textbf{Stopping }\\\textbf{Criterion}\end{tabular} & {\begin{tabular}[c]{@{}l@{}}\textbf{Validation}\\\textbf{Proportion}\end{tabular}} & {\textbf{Mean (TA)}} & {\textbf{~Std (TA)}} \\
\hline
\textbf{MLH} & \textbf{0} & \textbf{58.968} & \textbf{1.51} \\
\hline
\multirow{8}{*}{\begin{tabular}[c]{@{}l@{}}\textbf{Validation }\\\textbf{Accuracy}\end{tabular}} & 0.0001 & 53.948 & 3.48 \\
 & 0.0004 & 56.177 & 2.78 \\
 & 0.0016 & 57.407 & 2.12 \\
 & \textbf{0.0251} & \textbf{58.039} & 1.80 \\
 & 0.1 & 57.903 & 2.10 \\
 & 0.2 & 57.575 & 1.58 \\
 & 0.3 & 56.371 & 1.96 \\
 & 0.4 & 55.538 & 1.75 \\
\bottomrule
\end{tabular}
\caption{Mean and Standard deviation of Test Accuracy (TA) for 100 training runs with various validation proportions.\label{tab:valprop}}
\end{table}


In Early Stopping, the stopping criterion is monitored throughout the training procedure, and if a certain predefined condition involving the criterion is met, the training is stopped. Usually, the criterion used is the accuracy on the validation set.

If the data splits are not predefined, as they are in many competitions, the practitioner has to decide on the size of the splits. Early Stopping based on validation set criteria require a validation set to be split off from the training data, which, depending on the size, results in a significant reduction in the amount of available training data. The size of the validation set can also be seen as a hyperparameter. Larger validation sets can result in poorer models due to lower amounts of training data. On the other hand, smaller validation sets can result in inaccurate estimates of generalization performance, and the criteria being unreliable, leading to premature or late stopping. The optimal size of the validation set finds the best trade-off, as observed in Fig. \ref{fig:valprop}. But finding this optimal size of the validation set is non-trivial and requires multiple training runs.

For the experiments in this section, we use the CIFAR10 dataset, and a smaller AlexNet-like model without dropout to make sure the models overfit and hence make our observations concrete. We do a sweep of various validation set sizes and use \textit{Validation Accuracy} as the Early Stopping metric. For each setting, we train 100 such models for computing confidence intervals. Fig. \ref{fig:valprop} (Blue) illustrates the validation set size trade-off.

\begin{figure*}[htb]
    \centering
    \includegraphics[width=0.48\linewidth]{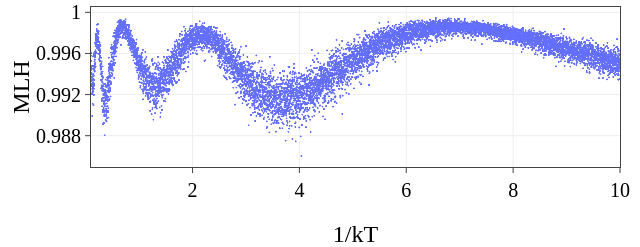}
    \includegraphics[width=0.48\linewidth]{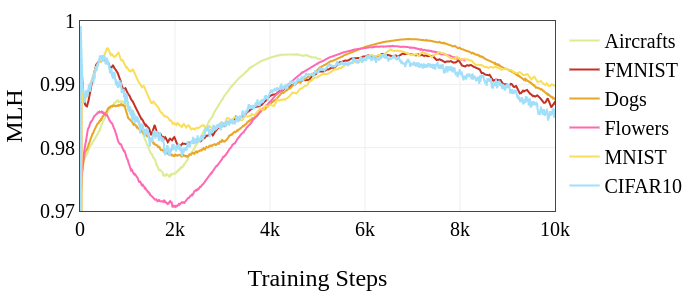}
    \caption{Left: MLH of Energy states at different values of Temperature $T$. Right: MLH of DenseNet121 weights on multiple datasets.}
    \label{fig:td-oscilation}
\end{figure*}

For one set of models, Fig. \ref{fig:valprop} (Red), we use MLH as the Early Stopping criterion, and include validation data for training. Fig. \ref{fig:valprop} shows that even if the practitioner knows the optimal validation set size beforehand, the mean test accuracy is significantly lower than when not using a validation set at all.

\section{MLH and Thermodynamics}\label{sec:thermodynamics}

In the previous sections, we used shallow Alexnet-Like networks that were prone to overfitting. Recent work has shown that larger and deeper neural network architectures are robust to overfitting \cite{gen}.

As observed in Fig. \ref{fig:td-oscilation} (Bottom), when we swap out the smaller AlexNet-like model with a larger DenseNet-121, we observe that the model never clearly overfits. As a result, MLH oscillates periodically; we note this observation on multiple datasets. 

We use this observation to present informal evidence that training NNs can be thought of as a thermodynamic process. We connect this oscillatory pattern of MLH to a contribution by \cite{physics} where they find that for systems following Boltzmann-Gibbs statistics, such as an ideal gas in a sealed chamber, their mantissa distribution of energy states of particles oscillates around BL with change in temperature. This is illustrated by Fig. \ref{fig:td-oscilation} (Top). Here, we run a simulation where we sample a large number of Energy states at a Temperature $T$ with the probability density function for an energy state $E$ from \cite{physics},
\begin{equation}
    f(E) = \frac{1}{k T} e^{-\frac{E}{k T}}
\end{equation}

Here, $k$ is the Boltzmann Constant. We compute MLH of energies at $1/kT=0.1$ to $1/kT=10$ at 10000 equally-spaced values. Fig. \ref{fig:td-oscilation} (Top) shows how MLH changes as a function of temperature $T$ which strikingly resembles Fig. \ref{fig:td-oscilation} (Bottom), where we plot 6 models trained on 6 different datasets, and compute MLH of their weights\footnote{Additional details are provided in the appendix}.

\begin{table}[tb]
\begin{tabular}{@{}ll@{}}
\toprule
\textbf{Deep Learning} & \textbf{Gas Chamber}      \\ \midrule
Weights                & Energy States             \\
Synaptic Connections   & Gas Particles             \\
SGD Steps              & Reciprocal of Temperature \\
Train Accuracy         & Heat Given                \\
MLH of Weights         & Heat Released             \\
\textbf{\begin{tabular}[c]{@{}l@{}}Train Accuracy  +\\ MLH\end{tabular}} & \textbf{Internal Energy} \\ \bottomrule
\end{tabular}
\caption{Analogies between Deep Learning and Thermodynamics}
\label{tab:analogy}
\end{table}

Furthermore, this motivates us to think about Temperature as an analogous to Gradient Descent iterations, and value of the weights analogous to the Energy states. In Table \ref{tab:analogy}, we list down the analogies drawn between Thermodynamics and Deep Learning. We can think of MLH of weights as the \textit{measure of stability}, i.e. higher MLH means the distribution of weights is more \textit{natural}.

Throughout this work, we assumed that MLH is a measure of model complexity, however, explaining why MLH contains this information would possibly also require us to answer why BL even emerges in the first place, which has remained unexplained since the phenomenon was discovered nearly two centuries ago.



\bibliography{aaai22}

\appendix
\section{Appendix}
\subsection{Variation of MLH across Layer Depth}

\begin{figure*}[htb]
\centering
\begin{minipage}{0.48\textwidth}
  \centering
\includegraphics[width=\textwidth]{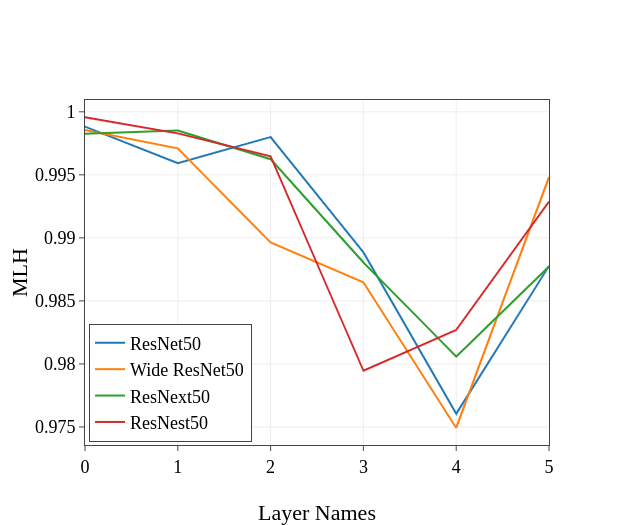}
\label{resnet}
\end{minipage}
\begin{minipage}{0.48\textwidth}
  \centering
\includegraphics[width=\textwidth]{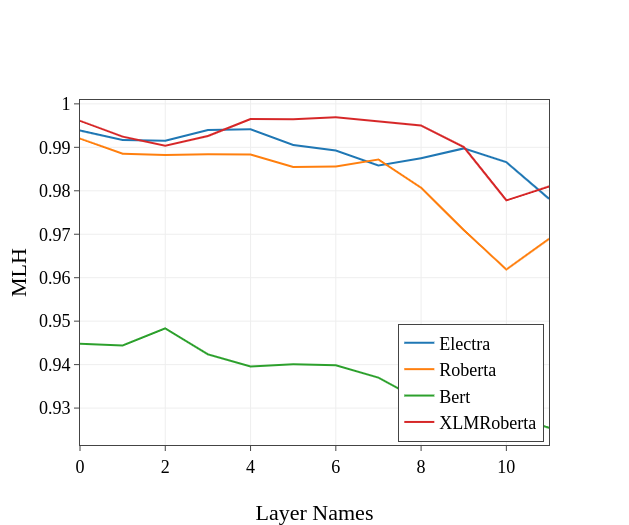}
\label{bert}
\end{minipage}

\caption{MLH across different layers of pretrained models.} \label{fig:layerwise}
\end{figure*}

In Fig. \ref{fig:layerwise}, we compute MLH for various pre-trained convolutional neural networks and Transformers and plot their layer-wise MLH. We see that generally, the MLH is high at the first layers, decreases gradually, but spikes again at the last layer. A similar trend is observed for Language Models (right).

\subsection{Effect of Data on MLH}

\begin{figure*}[htb]
\centering
\includegraphics[width=0.53\textwidth]{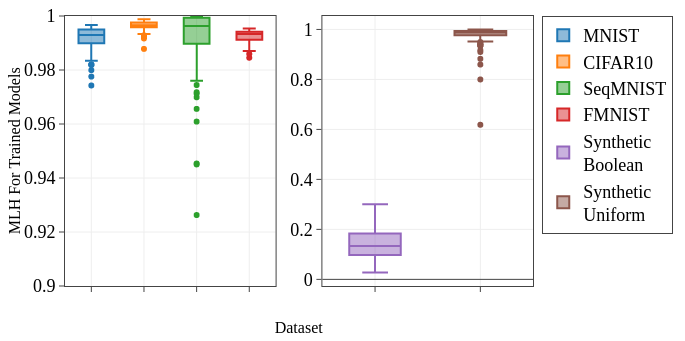}
\includegraphics[width=0.45\textwidth]{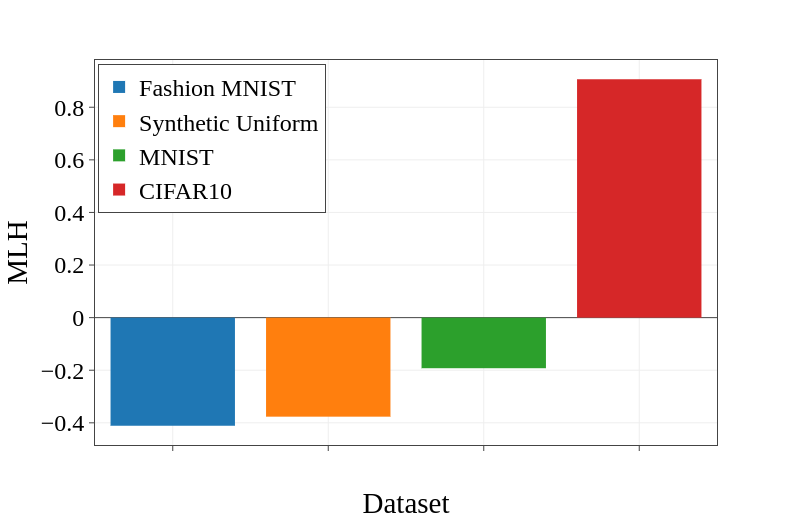}
\caption{MLH for Datasets and Networks Trained on them.}
\label{fig:dataset_analysis}
\end{figure*}

It is well known that RGB Images and their pixel values follow Benford's Law \cite{ganbenford}. We investigate whether the data distribution affects the Network Weights, causing them to match Benford's Law.

We investigate CIFAR10 \cite{krizhevsky2009learning}, MNIST \cite{lecun-mnisthandwrittendigit-2010}, FashionMNIST \cite{xiao2017fashion}, Sequential MNIST (SeqMNIST) \cite{goodfellow2013empirical}, two Synthetically Generated Datasets, and 100 Models trained\footnote{Experiments were conducted in an Ubuntu system, PyTorch 1.7, with a single NVIDIA RTX 2060 GPU, 16GB of RAM} on each of these datasets. We train a small LeNet-inspired network for MNIST and FashionMNIST, DenseNet121 for CIFAR10, LSTM \cite{hochreiter1997long} for SeqMNIST, and MLP for Synthetic Datasets.

For Synthetic-Boolean Dataset, We generate a random boolean function taking the $argmax$ over a randomly initialized Network with 64 boolean input features. Since the inputs are boolean, i.e., Base 2, Benford's Law for Base 2 is 1 \cite{10.2307/2160815}.

For Synthetic-Uniform Dataset, we do the same as above, but the inputs are drawn uniformly from the range $[0, 1]$.

Since the Scale Invariance Property of BL holds, we only divide by 255 for Image Datasets and do not normalize this experiment’s features.

As Observed in Figure\ref{fig:dataset_analysis}, CIFAR10 has positive MLH, and others have Negative MLH, whereas all of the models except for the ones trained on Synthetic-Boolean have $>$0.9 median MLH. Models trained on Synthetic-Boolean have a low but positive median MLH of about 0.2. We believe that MLH of the network is more related to feature variance than MLH of the Data.

\subsection{Preprocessing of Data}
For all experiments, we only divide by the maximum magnitude across features rather than normalizing them. This was to preserve the \textit{Scale Invariance} Property of Benford's Law.

\subsection{Model Architectures}
For all experiments in the paper, we use the following architectures and their respective datasets.
\begin{table}[tb]
\centering
\begin{tabular}{ll}
\toprule
\textbf{Dataset}           & \textbf{Architecture} \\ \hline
MNIST             & LeNet        \\ 
FashionMNIST      & LeNet        \\ 
CIFAR10           & DenseNet-121 \\ 
SequentialMNIST   & LSTM         \\ 
Synthetic-Uniform & MLP          \\ 
Synthetic-Boolean & MLP          \\
Oxford-Flowers  & DenseNet-121   \\ 
Stanford-Dogs  & DenseNet-121    \\ 
Aircrafts  & DenseNet-121        \\
\bottomrule
\end{tabular}
\caption{Model architectures used for training on corresponding datasets.}
\end{table}

The details of the models that have been used for the experiments are given below:
\begin{itemize}
    \item LeNet \cite{lecun1998gradient} architecture is a simple network with two(2) blocks of Strided Convolution-LeakyReLU Network followed by a fully-connected classification layer.
    \item DenseNet-121 \cite{huang2017densely} implementation is borrowed from this GitHub repository\footnote{\url{https://github.com/kuangliu/pytorch-cifar/blob/master/models/densenet.py}}.
    \item LSTM is a simple 2-layered RNN with Long-Short Term Memory \cite{lstm} followed by a fully-connected layer for classification.
    \item MLP is a 2 fully-connected network with Classification (Softmax) and Regression (Linear) output layers for Synthetic-Boolean and Synthetic Uniform Datasets.
\end{itemize}

\subsection{Hyperparameters}
Unless stated otherwise, all of the experiments use PyTorch's implementation of Adam \cite{kingma2014adam}, and PyTorch Lightning's  \cite{falcon2019pytorch} default values except the following wherever applicable.

\begin{table}[]
\centering
\begin{tabular}{ll}
\toprule
\textbf{Parameter}      & \textbf{Value} \\ \hline
Learning Rate           & 3e-3           \\
Early Stopping Patience & 15             \\
Validation Frequency    & 0.33           \\
Batch Size              & 64             \\ 
\bottomrule
\end{tabular}
\caption{Hyperparameters used during training}
\end{table}

\subsection{Data Splits}

For the datasets other than synthetically generated, we provide the Train/Validation/Test sets. All of them are randomly split unless PyTorch has Train/Test splits.

\begin{table}[htb]
\centering
\begin{tabular}{lccc}
\toprule
\textbf{Dataset} & \textbf{Train} & \textbf{Validation} & \textbf{Test} \\ \hline
MNIST            & 45000          & 5000                & 10000         \\ 
FashionMNIST     & 45000          & 5000                & 10000         \\ 
CIFAR10          & 45000          & 5000                & 10000         \\ 
SequentialMNIST  & 45000          & 5000                & 10000         \\
Synthetic-Boolean  & 6000          & 2000                & 2000         \\
Synthetic-Uniform  & 6000          & 2000                & 2000         \\ 
Oxford-Flowers  & 6149          & 2040                & 0         \\ 
Stanford-Dogs  & 12000          & 8580                & 0         \\ 
Aircrafts  & 6667          & 3333                & 0         \\ 
\bottomrule
\end{tabular}
\caption{Train, Validation and Test set splits for Datasets used.}
\end{table}

\subsection{Synthetic Data Generation}

We use synthetic data set for training a regression and a classification model. The synthetic data is generated using a single layered neural network that is randomly initialized. We describe the process of the generation process below. Both the data sets have an input vector length of 64 and these input vectors are also drawn randomly from Uniform $[0, 1]$ and Bernoulli for Synthetic-Uniform and Synthetic-Boolean respectively. We generate 10000 pairs of input-output. These are split in the ratio 60:20:20 for Train, Validation and Test sets respectively.

\subsection{Regression data set}

$x_i \in \mathbf{R}^{64}$ is an input vector sampled from $Uniform [0, 1]$. The target label $y_i \in \mathbf{R}$ is obtained by feeding $x_i$ as input to a randomly initialized single layered linear kernel with one output unit, $f_\theta$. 
\begin{equation}
    y_i = f_\theta(x_i), x_i \in Uniform[0, 1] 
\end{equation}

\subsection{Classification data set}

$x_i \in \mathbf{R}^{64}$ is an input vector sampled from $Bernoulli [0, 1]$. The target label $y_i \in \mathbf{R}$ is obtained by feeding $x_i$ as input to a randomly initialized single layered linear kernel with two output units and argmax-ed over the output vector. 
\begin{equation}
    y_i = argmax[f_\theta(x_i)], x_i \in Bernoulli[0, 1] 
\end{equation}

\begin{figure}[tb]
\begin{verbatim}
benford = np.array([30.1, 17.6, 12.5, 9.7,
                    7.9, 6.7, 5.8, 5.1, 4.6]
            ) / 100

def mlh(bin_percent):
    return scipy.stats.pearsonr(
                benford, 
                bin_percent[1:]
    )[0]

def bincount(tensor):
    counts = torch.zeros(10)
    for i in range(10):
        counts[i] = torch.count_nonzero(
            tensor == i
        )
    return counts
    
@torch.no_grad()
def bin_percent(tensor):
    tensor = tensor.abs() * 1e10
    long_tensor = torch.log10(tensor).long()
    tensor = tensor // 10 ** long_tensor
    tensor = bincount(tensor.long())
    return tensor / tensor.sum()
\end{verbatim}
\caption{PyTorch code for computing MLH}
    \label{fig:code}
\end{figure}

\subsection{PyTorch Code for Computing MLH}

In the code provided in Fig. \ref{fig:code}, the vector $benford$ is the distribution defined by Benford's Law, $bin\_percent$ is a function that takes a tensor as input and returns the distribution of digits in it. $bincount$ is a function that takes a flattened tensor of integers $\in[0, 9]$ and returns a vector of frequencies. MLH is the function to compute the proposed MLH score.

\end{document}